%% file: main.tex
\documentclass[10pt,twocolumn,letterpaper]{article}

\usepackage[pagenumbers]{cvpr} % To force page numbers, e.g. for an arXiv version

\input{preamble}
\definecolor{cvprblue}{rgb}{0.21,0.49,0.74}
\usepackage[pagebackref,breaklinks,colorlinks,allcolors=cvprblue]{hyperref}
\usepackage{algorithm}
\usepackage{listings}
\usepackage{xcolor}
\def\paperID{} % *** Enter the Paper ID here
\def\confName{CVPR}
\def\confYear{2026}

\title{Multi-Patch Global-to-Local Transformer Architecture For Efficient \\Flow Matching and Diffusion Model}

\author{Quan Dao\\
Rutgers University\\
{\tt\small quan.dao@rutgers.edu}
\and
Dimitris Metaxas\\
Rutgers University\\
{\tt\small dnm@cs.rutgers.edu}
}

\begin{document}
\maketitle
\begin{figure*}[ht!]
    \centering
    \includegraphics[width=1\linewidth]{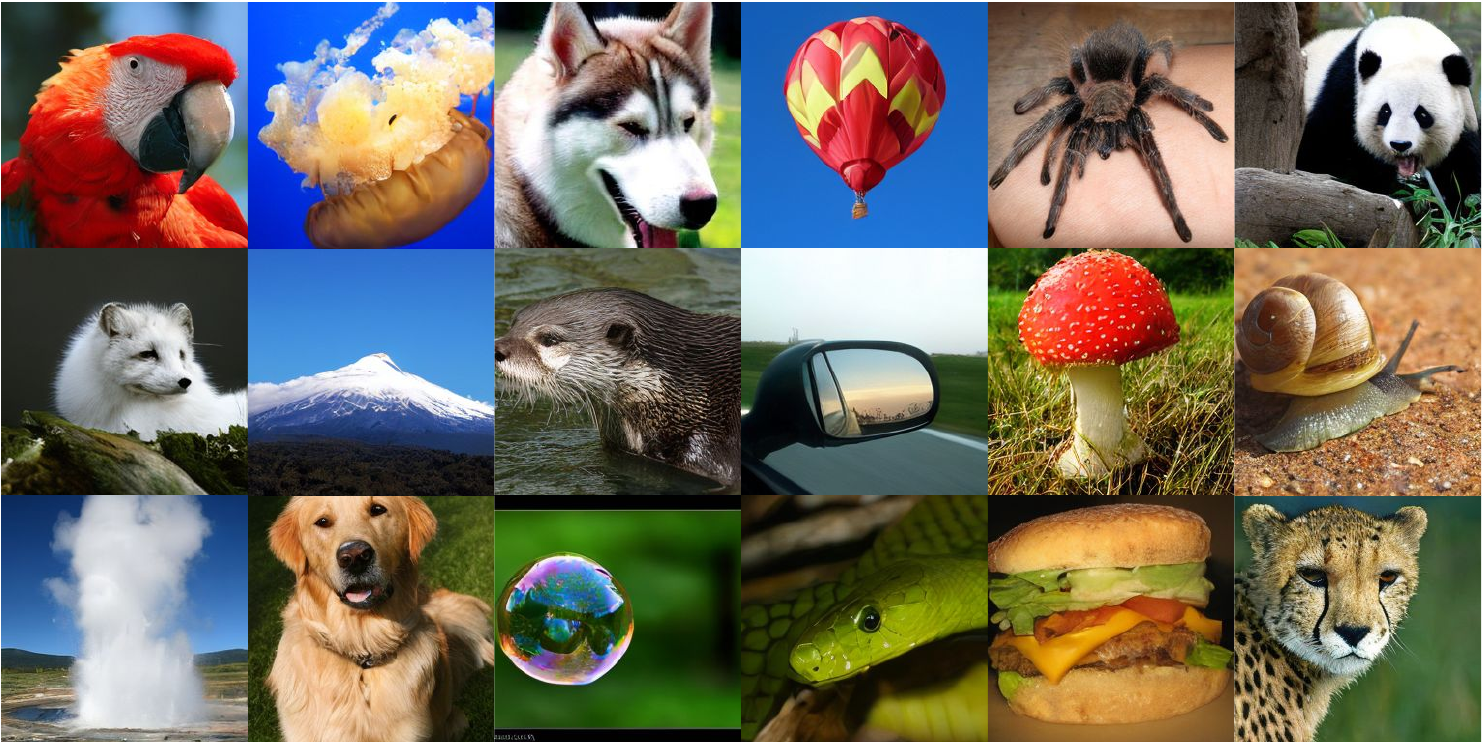}
    \caption{The generated samples from MPDiT-XL with the cfg-scale $w=3$ at epoch 160.}
    \label{fig:teaser}
\end{figure*}
\input{sec/0_abstract}    
\input{sec/1_intro}

\input{sec/2_related}

\input{sec/3_method}

\input{sec/4_exp}
\input{sec/5_conclusion}

% \clearpage
{
    \small
    \bibliographystyle{ieeenat_fullname}
    \bibliography{main}
}
\input{sec/X_suppl}
\end{document}

%% file: sec/0_abstract.tex
\begin{abstract}
Transformer architectures, particularly Diffusion Transformers (DiTs), have become widely used in diffusion and flow-matching models due to their strong performance compared to convolutional UNets. However, the isotropic design of DiTs processes the same number of patchified tokens in every block, leading to relatively heavy computation during training process. In this work, we introduce a multi-patch transformer design in which early blocks operate on larger patches to capture coarse global context, while later blocks use smaller patches to refine local details. This hierarchical design could reduces computational cost by up to 50\% in GFLOPs while achieving good generative performance. In addition, we also propose improved designs for time and class embeddings that accelerate training convergence. Extensive experiments on the ImageNet dataset demonstrate the effectiveness of our architectural choices. Code is released at \url{https://github.com/quandao10/MPDiT}
\end{abstract}

%% file: sec/1_intro.tex
\section{Introduction}
\label{sec:intro}

Diffusion models \cite{ho2020denoising, song2020score, lipman2022flow, dhariwal2021diffusion} have emerged as a leading class of generative models, surpassing generative adversarial networks \cite{goodfellow2014generative}, normalizing flows \cite{dinh2016density, kingma2018glow, zhai2024normalizing}, and autoregressive models \cite{oord2016pixel, tian2024visual, van2016conditional} in many vision tasks. Compared to GANs \cite{goodfellow2014generative}, diffusion models \cite{ho2020denoising} are generally easier to train and avoid issues such as instability and mode collapse. In 2D image generation, diffusion-based approaches have demonstrated strong performance in text-to-image synthesis \cite{rombach2022high}, enabling downstream applications such as personalization \cite{ruiz2023dreambooth, ye2023ip, van2023anti}, image editing\cite{meng2021sdedit, huberman2024edit, nichol2021glide, he2024dice, dao2025discrete, pham2025autoedit}, and text-to-3D generation\cite{poole2022dreamfusion, wang2023prolificdreamer}. In 3D and video domains, text-to-video and image-to-video diffusion models \cite{ho2022video, yang2024cogvideox, singer2022make} have also shown promising results, producing high-quality videos that align well with textual or visual conditions. As diffusion models continue to achieve state-of-the-art results across modalities, recent research has increasingly focused on improving their efficiency \cite{yu2025repa, hang2023efficient, xie2024sana, xie2025sana} while maintaining generation quality, aiming to make diffusion models faster and more practical for large-scale deployment.

Despite their strong performance in visual generation tasks, diffusion models remain computationally expensive to train and sample. To address the high sampling cost, recent research has explored two main directions: higher-order numerical solvers \cite{lu2022dpm, zhang2022fast} and diffusion distillation \cite{meng2023distillation, yin2024improved, salimans2022progressive, yin2024one, nguyen2024swiftbrush}. Sampling efficiency has been extensively studied, and recent distillation methods \cite{yin2024improved, yin2024one, dao2025improved, dao2025self, zhang2025flow} can achieve performance comparable to their teacher diffusion models. For training efficiency, recent works have shifted from pixel diffusion \cite{ho2020denoising} to latent diffusion \cite{rombach2022high, dao2023flow}, which is considerably faster. However, training diffusion models in latent space still demands substantial computational resources. To mitigate this, ongoing research focuses on developing more compact and semantically meaningful variational autoencoders (VAEs) \cite{chen2025dc, chen2024deep, kouzelis2025eq, yao2025reconstruction} that provide efficient representations and enable faster convergence. 
Beyond improving VAE design, two other orthogonal directions have been explored: objective alignment \cite{yu2025repa} and architectural design \cite{teng2024dim, wang2025lit, zhu2025dig, ai2025dico, tian2025dic}. Objective alignment methods, such as REPA \cite{yu2025repa}, leverage pretrained self-supervised models like DINO \cite{oquab2023dinov2, caron2021emerging} to regularize diffusion features. This approach significantly enhances training stability and sample quality compared to using the standard diffusion objective. 

From an architectural perspective, diffusion models initially adopted UNet backbones \cite{song2020score, ho2020denoising, dao2024high} but have increasingly transitioned to transformer designs \cite{bao2023all, peebles2023scalable} due to their strong scalability. Similar to advances in visual perception and large language models, transformer architectures demonstrate robust performance when scaled to large models for text-to-image \cite{chen2023pixartalpha} and text-to-video generation \cite{yang2024cogvideox, wan2025wan, sora_openai_2025}. Recent works have introduced efficient transformer variants, such as linear attention \cite{xie2024sana, wang2025lit, zhu2025dig}, to alleviate the quadratic complexity of full attention and reduce memory and computation costs. However, linear attention often struggles to capture long-range dependencies, trading some performance for efficiency. State-space architectures\cite{gu2024mamba, gu2021efficiently}, such as Mamba diffusion models \cite{phung2024dimsum, teng2024dim, yan2024diffusion}, have also been explored. Yet, they offer limited benefits for diffusion tasks, as their advantages typically emerge with very long token sequences, whereas most latent-space diffusion models involve fewer than a thousand tokens. Meanwhile, convolution-based architectures \cite{ai2025dico, tian2025dic} have recently been revisited, showing competitive or even superior performance with reduced training time.

In this paper, we revisit the transformer architecture design for diffusion models to reduce both parameter count and computational cost (GFLOPs) while preserving high generative performance. We validate our proposed design through extensive experiments on the ImageNet dataset. We introduce MPDiT, a global-to-local diffusion transformer architecture that processes information at multiple patch scales. In the early stage, the model uses large-patch tokenization in the first 
$N-k$ transformer blocks to efficiently capture global contextual information with a smaller number of tokens. In the later stage, an upsample module expands these large-patch tokens into a greater number of small-patch tokens. The resulting fine-grained tokens are then processed by the final k transformer blocks, which focus on refining local details and improving the visual quality of the generated image. We find that using only a small number of refinement blocks $(k=4 \rightarrow 6)$ is sufficient for high quality image synthesis. This design is conceptually inspired by the idea of global-local attention \cite{hatamizadeh2023global, liu2021swin, yang2021focal}, but rather than applying it within each transformer block, which could hurt the performance while only reducing a negligible computation budget, we apply it at the architectural level, achieving greater efficiency and better performance under the same computational budget.
In addition, we reexamine the time embedding mechanism. Instead of the conventional linear time embedding on sinusoidal embedding of time $t$ \cite{nichol2021glide, peebles2023scalable}, we propose a fourier neural operator time embedding which is motivated from Neural Operator layer \cite{li2020fourier}, which captures richer temporal dependencies and yields approximately a 4 points FID improvement. 
For class conditioning, rather than using the AdaIN modulation adopted in DiT \cite{peebles2023scalable}, we follow the UViT \cite{bao2023all} approach that add prefix class token before the input token sequence. We further extend this idea by using multiple class tokens instead of a single token, which improves convergence under limited training budgets. This suggests that representing class information with multiple tokens allows the model to capture richer semantic structure, leading to more effective interaction between class and spatial image tokens. 
% \quan{might need wrap performance here}

In summary, our main contributions are as follows:

\begin{enumerate}
    \item \textbf{Global-to-Local transformer architecture:} We propose a hierarchical transformer architecture that processes visual information in a coarse-to-fine manner. The model first operates on large-patch tokens to efficiently capture global context, then progressively upsamples them into small-patch tokens for fine-grained refinement. This architecture embodies the idea of global-local attention, but applies it at the network level rather than within attention layer, achieving a better balance between efficiency and generation quality.
    \item \textbf{Revisit diffusion transformer components:} We revisit conditioning modules of diffusion transformers, including time and class embeddings. For \textbf{time embedding}, we replace the conventional linear embedding with a \textbf{FNO embedding} to capture smoother transitions across timesteps and provide richer temporal representations. For \textbf{class conditioning}, we introduce \textbf{multi-token embedding}, enabling  amore expressive representation of condition and improving training convergence.
    \item \textbf{Comprehensive evaluation:} We perform extensive experiments on the ImageNet dataset to validate the effectiveness of our architectural design.
\end{enumerate}

%% file: sec/2_related.tex
\section{Related Works}
\label{sec:related}

\subsection{Efficient Training Strategy } \label{sec:related:strat}
Diffusion models \cite{song2020score, lipman2022flow, ho2020denoising, dhariwal2021diffusion} have achieved state-of-the-art performance in visual generation tasks, surpassing other generative approaches such as GANs \cite{goodfellow2014generative} and autoregressive models \cite{van2016conditional, oord2016pixel, zhai2024normalizing}. They are also preferred for their stable training dynamics, which eliminate the need for delicate hyperparameter tuning required by GANs. However, diffusion training remains computationally demanding due to its slow convergence. Early efforts to improve training efficiency primarily focused on loss reweighting \cite{hang2023efficient} and timestep sampling strategies \cite{esser2024scaling}. These approaches treat diffusion or flow-matching training as a multi-task learning problem, emphasizing mid-range timesteps where the signal-to-noise balance is optimal. By prioritizing these timesteps, such methods effectively reduce gradient variance and accelerate convergence without compromising model quality. 

Recently, REPA \cite{yu2025repa, tian2025u} introduced a method to align diffusion features with pretrained representations extracted from DINO \cite{oquab2023dinov2, caron2021emerging}, leveraging the strong semantic structure of self-supervised features to significantly accelerate training. Diffuse and Disperse \cite{wang2025diffuse} proposed a dispersive loss in feature space, serving as a regularizer that integrates self-supervised learning principles into the diffusion training process. Another approach, $\Delta$FM \cite{stoica2025contrastive}, introduced a contrastive regularization objective that pushes the predicted velocity away from the ground-truth velocity of mismatched pairs, thereby improving feature discrimination and representation quality.

Another line of research focuses on designing VAEs \cite{yao2025reconstruction, kouzelis2025eq, chen2025dc, chen2024deep} that aggressively compress the spatial dimension while increasing the channel dimension, achieving more compact latent representations. Methods such as DC-VAE \cite{chen2024deep} and DC-AE 1.5 \cite{chen2025dc} proposes deep compressed autoencoder that enable deep spatial compression, effectively reducing the number of latent tokens and improving overall training efficiency. In this paper, we instead focus on the backbone design of diffusion and flow matching models to reduce both training time and inference cost while maintaining strong generative performance. In \cref{sec:related:backbone}, we summarize various architectural approaches aimed at improving the efficiency and convergence of diffusion models.

\subsection{Diffusion Backbone Design}\label{sec:related:backbone}
The early architectures for diffusion and score-based generative models primarily adopted a UNet backbone \cite{ho2020denoising, song2020score}, motivated by the fact that both the input and output share the same spatial resolution. Subsequent works such as UViT \cite{bao2023all} and DiT \cite{peebles2023scalable} introduced transformer-based architectures, achieving notable performance gains over UNet models. These transformer backbones have demonstrated strong scalability, achieving high-quality results on large-scale tasks such as text-to-image and text-to-video generation.
Despite their success, diffusion transformers \cite{bao2023all, peebles2023scalable} remain computationally expensive, requiring extensive training budgets and high-end GPUs with large memory and fast processing speed. A major source of this inefficiency arises from the full attention layers, which scale quadratically with the number of tokens. To address this issue, SANA \cite{xie2024sana} introduced ReLU-based linear attention layers to reduce memory usage and computational overhead. However, this approach often results in a noticeable performance drop compared to full attention as shown in LIT \cite{wang2025lit}. LIT \cite{wang2025lit} further improves upon ReLU linear attention by proposing an enhanced linear attention formulation, yet it still requires initialization from a pretrained full-attention model to achieve competitive results, rather than being trained entirely from scratch. In addition, DiG \cite{zhu2025dig} introduces a gated linear attention mechanism that attains competitive results. 

Recently, state-space architectures\cite{gu2021efficiently}, particularly Mamba\cite{gu2024mamba}, have been explored as potential replacements for transformers in diffusion models. Several studies, including DiM \cite{teng2024dim}, DifuSSM \cite{yan2024diffusion}, and Zigma\cite{hu2024zigma}, adopt Mamba-based designs, while DiMSUM \cite{phung2024dimsum} introduces a hybrid Mamba–Attention architecture. However, these approaches achieve performance comparable to transformer-based diffusion models, with limited efficiency gains. This is primarily because Mamba exhibits substantial speed advantages only when processing very long token sequences (typically exceeding 1k tokens), whereas latent diffusion models \cite{rombach2022high} usually operate with fewer than 1k tokens. In parallel, DiCo \cite{ai2025dico} and DiC\cite{tian2025dic} revisit convolution-based architectures, demonstrating strong performance with significantly lower GFLOPs. Another line of work focuses on token reduction through masked modeling, as explored in MaskDiT \cite{Zheng2024MaskDiT}. The MaskDiT adopts an encoder-decoder framework, where the encoder processes the visible (unmasked) tokens and the decoder reconstructs the masked tokens. However, MaskDiT exhibits a notable performance degradation when a large fraction of tokens is masked and generally requires additional fine-tuning with full tokens to recover generation quality.

Our work is inspired by the concept of global–local attention \cite{liu2021swin, yang2021focal}, which aims to balance efficiency and representation capacity. However, merely substituting global-local attention \cite{crowson2024scalable} could fail to achieve meaningful efficiency gains and typically incur a reduction in performance. To address this, we extend the global–local modeling concept to the architectural level rather than individual attention layers. In addition, we introduce a Fourier Neural Operator (FNO) \cite{li2020fourier} time embedding to learn more expressive temporal representations, and a multi-token class embedding strategy to enhance conditional modeling. Together, these components lead to faster training convergence and maintain good generation quality.

%% file: sec/3_method.tex
\section{Method}\label{sec:method}

\begin{figure*}
    \centering
    \includegraphics[width=\linewidth]{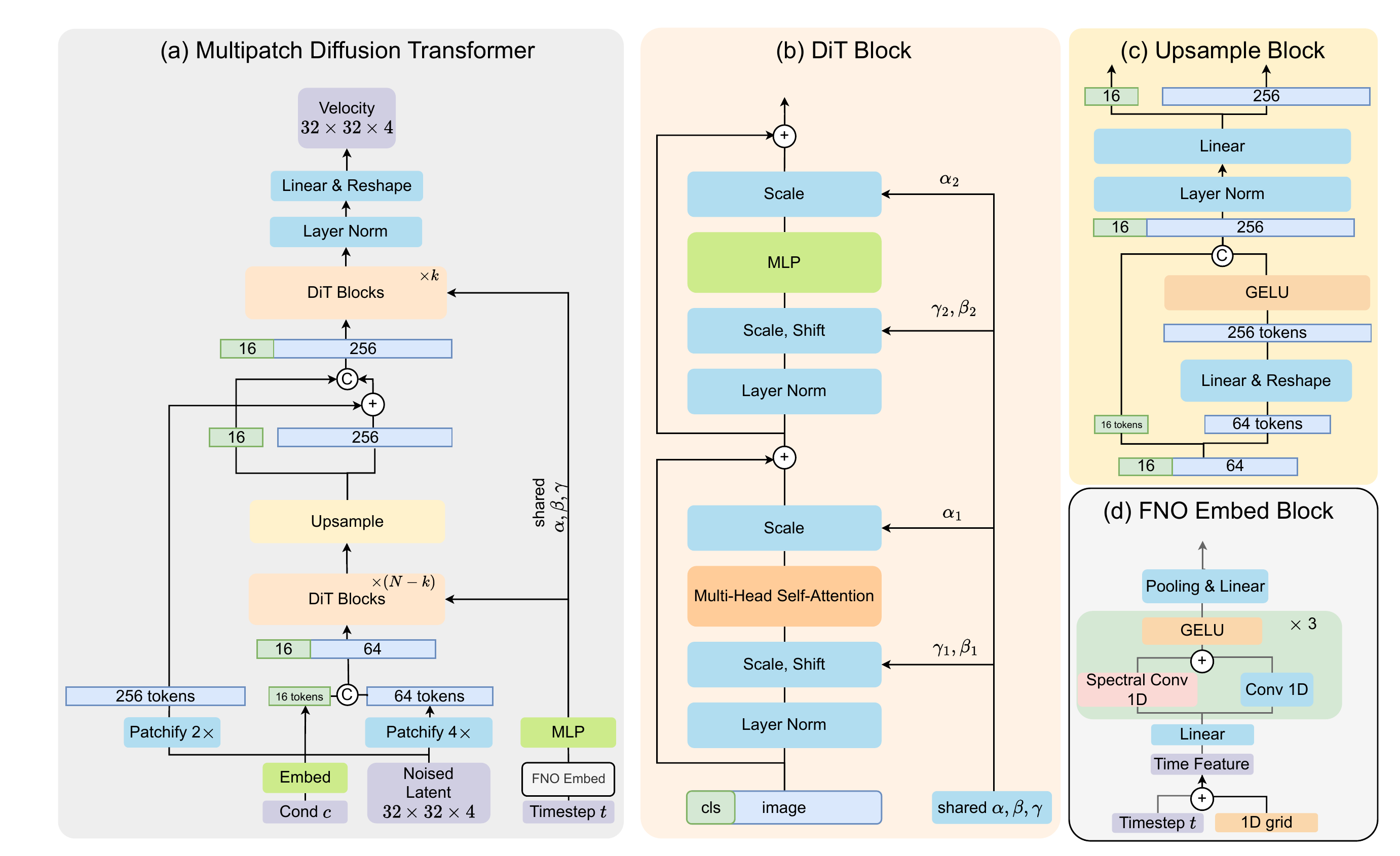}
    \caption{\textbf{Architecture of MPDiT}, which consists of (a) the Global-Local MultiPatch Diffusion Transformer, (b) DiT Block with shared time embedding, (c) The Upsample Module and (d) The FNO Time Embedding}
    \label{fig:sysfig}
\end{figure*}

In this section, we introduce our global-to-local transformer architecture and the proposed embedding modules for time and class. We first outline the latent flow-matching training pipeline in \cref{sec:sub:prelim}. Then, in \cref{sec:sub:mpdit}, we describe the details of the global-to-local transformer design MPDiT. To capture global information efficiently, the model employs a large-patch embedding module that reduces the number of input tokens. The initial transformer blocks operate on these large-patch tokens to model global context. The token sequence is then upsampled to a fine-grained smaller-patch tokens, and the subsequent transformer blocks process these tokens to capture local image details. Finally, in \cref{sec:sub:emb}, we present our FNO-based time embedding and multi-token class embedding modules, which significantly enhance model performance and accelerate convergence.

\subsection{Preliminaries}\label{sec:sub:prelim}
Given a training dataset $\mathcal{D}$ containing images $x \in \mathbb{R}^{H \times W \times 3}$. , we denote the encoder and decoder of the VAE as $\mathcal{E}$ and $\mathcal{D}$, respectively. In latent diffusion or flow matching, the encoder $\mathcal{E}$ maps an image $x$ from pixel space to a latent representation $z = \mathcal{E}(x) \in \mathbb{R}^{h \times w \times d}$, where $h = H/r$, $w = H/r$, $r$ is the VAE compression ratio, and $d$ is the latent channel dimension. The goal of latent flow matching \cite{lipman2022flow} is to learn a velocity prediction model $f_\theta$that estimates the target velocity given the noisy latent input, timestep, and optional class condition. The corresponding training objective is defined as:

\begin{equation}
    L_{FM} = \sum_{z, t, n, \epsilon}||f_\theta(z_t, t, c) - (n-z)||_2^2
\end{equation}
where noise $n \sim \mathcal{N}(0, I_{h\times w \times d})$, noisy latent $z_t = (1-t)z + tn$. The time $t$ is sampled from range $[0, 1]$ and $c$ represents the conditioning signal. For example, a class index in class-to-image generation or a text caption in text-to-image generation.

After training the model $f_\theta$, the generative process can be performed by integrating backward from a random Gaussian noise latent to a clean latent representation $z$ using numerical solvers such as Euler, Heun, or Dopri5. Once the clean latent $z$ is obtained, the decoder $\mathcal{D}$ of the VAE is applied to reconstruct the corresponding image: $x = \mathcal{D}(z)$.

\subsection{Multi-patch Transformer}\label{sec:sub:mpdit}

\paragraph{Global-to-Local Design:}
We aim to develop an efficient transformer architecture for training diffusion and flow-matching models. A common strategy to reduce the time and memory complexity of transformer-based models is \textbf{token reduction}. This can generally be achieved in two ways: (1) by introducing global-local attention, which limits attention computation to selected regions or token subsets, or (2) by applying masked token modeling during training, where only a subset of tokens is processed while the remaining ones are reconstructed.

In visual perception tasks, several studies have explored global-local attention mechanisms to improve transformer efficiency, including Swin Attention \cite{liu2021swin}, Focal Attention \cite{yang2021focal}, and Global-Context (GC) Attention \cite{hatamizadeh2023global} variants. In these approaches, global attention is applied across global tokens representing different contextual windows, often obtain by using pooling operations, while local attention is restricted to local tokens within each contextual local window. This design reduces the number of tokens participating in self-attention, theoretically improving convergence and efficiency. However, these attention's performance often lags behind self-attention on full sequence of tokens, with limited practical efficiency gains \cite{crowson2024scalable}. The main source of inefficiency could arises from the repeated supporting operations (e.g., reshaping, pooling, and window partitioning) that must be executed in every transformer block and the fact that the latent diffusion often does not model large number of image tokens (less than 1k). 
%These operations accumulate substantial computational overhead when applied across a deep transformer architecture, diminishing the expected efficiency benefits. \quan{plot 2 type of global-local attention here and their performance with Gflops - refer to HDiT}. 

In diffusion models, MaskDiT \cite{Zheng2024MaskDiT} investigate \textbf{token reduction} through masking techniques and propose an encoder-decoder architecture to perform both denoising and reconstruction prediction. However, MaskDiT exhibits a significant performance degradation when a large masking ratio (e.g., 75\%) is applied. Specifically, MaskDiT-XL/2 \cite{Zheng2024MaskDiT} reports an FID of around $100$ at a $75\%$ mask ratio. In comparison, DiT-XL/4 \cite{peebles2023scalable}, which process the similar number of tokens, achieves a much stronger result of approximately 40 FID, while being more computationally efficient since it does not need additional decoder layers to reconstruct the full token set. This discrepancy arises because heavy masking causes each training example to learn only partial relationships among the remaining unmasked tokens, resulting in poor modeling of both global and local information. In contrast, DiT-XL/4 focuses on modeling global information using large-patch tokens, which, although lacking local detail, effectively captures the overall structure of the image. This observation suggests that adding refinement transformer blocks to enhance local representation could further improve the performance of such architectures.

From the above observations, we propose a simple yet efficient transformer architecture called the \textbf{Multi-Patch Diffusion Transformer (MPDiT)}. Given a standard DiT architecture with $N$ transformer blocks, as illustrated in \cref{fig:sysfig}(a), we consider the ImageNet-256 setting with a VAE encoder whose latent representation has shape $(32, 32, 4)$, corresponding to $1024$ tokens. 
A standard DiT applies a \textbf{$\texttt{PatchEmbed}_{p=2}(z_t)$}, where $p$ denotes the patch size, reducing the token count to $256$ for more efficient training. 
To further improve efficiency, MPDiT replaces the standard $p=2$ patch embedding with a larger patch size $p=4$ for the first $(N-k)$ transformer blocks. 
Increasing the patch size reduces the token count from $256$ to $64$, meaning these early blocks operate on only $25\%$ of the tokens used in a standard DiT. 
This coarse representation is sufficient for modeling global information and leads directly to a substantial reduction in GFLOPs, since self-attention scales quadratically with the number of tokens.

We then design an \textbf{Upsample Block} to expand the token sequence from $64$ to $256$ tokens, corresponding to an effective patch size of $p=2$. A skip connection from the $256$ tokens produced by the \textbf{$\texttt{PatchEmbed}_{p=2}(z_t)$} module is added directly to the upsampled tokens, ensuring that fine-grained details are preserved. The resulting $256$ tokens, containing both global information from the first $N-k$ blocks and original spatial features, are then processed by only the last $k$ DiT blocks to refine local details. In practice, we find that a small number of refinement blocks ($k=4 \rightarrow 6$) is sufficient to maintain strong generative performance while achieving a significant reduction in computational cost. 
Since the first $(N - k)$ blocks operate on only $\tfrac{1}{16}$ of the attention tokens, MPDiT could achieves up to a \textbf{50\% reduction in GFLOPs} for $\text{MPDiT-XL}_{k=6}$
% To illustrate the computational advantage, we compare the theoretical complexity of MPDiT and the standard DiT. 

For high resolution $512^2$, MPDiT could be more efficient by extending to a three-level patch hierarchy with $p \in \{8,4,2\}$. 
The first $(N - r_1 - r_2)$ blocks operate on 64 tokens ($p=8$), the next $r_2$ blocks operate on 256 tokens ($p=4$), and the final $r_1$ blocks operate on 1024 tokens ($p=2$). 
This yields a coarse-to-mid-to-fine representation that scales efficiently to larger spatial resolutions.
For ImageNet-256 latents, we find that a two-level patch hierarchy $\{4,2\}$ is sufficient to maintain strong performance while providing significant computational savings.

\paragraph{Upsample Block:} The Upsample Block plays a crucial role in MPDiT. 
A well-designed upsampling module is essential for ensuring that MPDiT matches the performance of a full-token DiT. 
As illustrated in \cref{fig:sysfig}(c), the input token sequence is first separated into class tokens and image tokens. 
The image tokens are then upsampled using a linear projection followed by a pixel-unshuffle operation, similar to the unpatchify step in DiT. 
This increases the sequence length from $64$ tokens to $256$ tokens (a $\times 4$ expansion in spatial token count).
The resulting $256$ image tokens pass through a GELU activation and are concatenated with the class tokens before entering a lightweight linear refinement block. 
Since the first $(N-k)$ blocks model interactions between the class tokens and the $64$ global tokens, the upsample operation may introduce misalignment between class and image tokens. 
To correct this mismatch, we include an additional linear layer that re-establishes a relationship between class and image tokens. 
LayerNorm is applied before the refinement step to stabilize gradients, and the GELU activation provides a rich non-linear mapping necessary for recovering fine-grained spatial details.

\paragraph{Other Details:}  We follow PixArt-$\alpha$ \cite{chen2023pixartalpha} and share the time-embedding module across all transformer blocks, which reduces the number of parameters and improves memory efficiency. 
This modification, however, yields only a small reduction in GFLOPs since the main computational bottleneck lies in the attention layers. 
In addition, we decouple the time and class conditioning: rather than injecting both signals through an AdaIN layer, 
we adopt learnable class tokens concatenated as a prefix to the image tokens, following the UViT \cite{bao2023all}. Note that other specification in DiT block remains the same (i.e \texttt{LayerNorm} and \texttt{AbsolutePositionEncoding})
% This design removes the burden of AdaIN modulation and provides a cleaner and more efficient conditioning mechanism.

\subsection{Revisiting Time and Class Embedding}\label{sec:sub:emb}
In addition to the MPDiT design, we also revisit the conditioning mechanisms for time and class information. 
% First, we introduce an FNO-based time embedding, which enables the model to learn smooth and expressive temporal features and leads to improved generative performance. 
% Second, we propose a multi-token class representation: instead of using a single class token, we employ multiple learnable class tokens, providing a richer and more distributed encoding of the class information. 
% This multi-token conditioning supplies stronger contextual signals to the image tokens and improves overall model quality.
\paragraph{Time Embedding Block.}
Although time is a fundamental component of diffusion and flow-matching models, the commonly used time-embedding design remains simple: the timestep is encoded with sinusoidal frequencies and then processed by a small MLP \cite{peebles2023scalable, nichol2021glide}. 
The time embedding design has received limited investigation, and it is unclear whether it is optimal for modeling the continuous dynamics of diffusion trajectories. 
Motivated by Neural Operator \cite{li2020fourier}, which are designed to learn smooth functions and physical dynamics, we introduce an FNO  time embedding that better captures the continuous flow field \cite{lipman2022flow} with underlying SDE and ODE equation.
The structure of our FNO time embedding is illustrated in \cref{fig:sysfig}(d). 
We first construct a 1D grid of 32 evenly spaced positions using \texttt{linspace(-1, 1, 32)}, which is added to the scalar timestep $t$ to form a 1D time signal and each signal now has single channel. 
This signal's channel is then lifted from dimension $1$ to dimention $32$ by a linear projection (the dimension $32$ is chosen heuristicly from the set  \{16, 32, 64, and 128\}. Dimension $128$ is unstable and $32, 64$ gives the best performance). 
Next, we apply three \texttt{MixedFNO} blocks, each consisting of a mixed \texttt{SpectralConv1D} and \texttt{Conv1D}, enabling the embedding to learn smooth and expressive temporal structure. 
The resulting feature is then average-pooled and passed through a final linear layer to project the feature from $d$ channels to the model's embedding dimension. The pseudo-code for the FNO time embedding module is provided in the Appendix.

\paragraph{Class Embedding Block} Traditional models often use a single token to represent the class label, which results in an overly dense class embedding and may slow training convergence. 
To address this, we introduce a \textbf{multi-token class embedding}, where each class is represented by several learnable tokens instead of a single vector. 
Let $c \in \{1,\dots,C\}$ be the class index and $m$ the number of class tokens, and 
$D$ the hidden dimension of the transformer (i.e., the model embedding dimension) 
We learn a class embedding matrix 
$
E_{\text{cls}} \in \mathbb{R}^{C \times (mD)},
$
and obtain the class tokens by reshaping the corresponding row:
\[
T_{\text{cls}} = \mathrm{reshape}\!\left(E_{\text{cls}}[c],\, (m, D)\right).
\]
These $m$ class tokens are prepended to the image tokens, providing a more distributed and expressive class representation. 
With this multi-token formulation, we observe consistently improved performance and faster convergence, as demonstrated in our ablation results.

%% file: sec/4_exp.tex
\section{Experiment}\label{sec:exp}

In this section, we evaluate our method on the ImageNet dataset \cite{deng2009imagenet} for class-conditional image generation at a resolution of $256 \times 256$. ImageNet contains 1,281,167 training images across 1,000 classes. Following standard practice \cite{dhariwal2021diffusion}, we report Frechet Inception Distance (FID)\cite{heusel2017gans}, Inception Score (IS)\cite{salimans2016improved}, Precision (Pre) \cite{kynkaanniemi2019improved}, and Recall (Rec) \cite{kynkaanniemi2019improved} to assess generative quality on 50K samples using Euler Solver with 250 steps. To measure efficiency, we report GFLOPs and number of parameters (Params).

\noindent\textbf{Training Details.} All experiments are conducted on a single A100 node with 8 GPUs (40GB each).
We use a fixed learning rate of $2 \times 10^{-4}$ with a total batch size of 1024, 
which is equivalent to a learning rate of $1 \times 10^{-4}$ with batch size 256.
We apply EMA with a decay rate of 0.9999 and report all results using the EMA checkpoint.
% All models are trained with mixed-precision \texttt{bf16}. For all variant of MPDiT, we train for 240 epoches. With a single node A100 40GB, the estimate training time for B size model is 14 hours (112 A100 hours) and XL size model is 44 hours (352 A100 hours). 

\subsection{Main Results} \label{sec:exp:main}

% In the main experiments, the configuration of MPDiT is provided in \cref{tab:mpdit_config}. For the experiment on ImageNet $256 \times 256$ with two variants MPDiT-B and MPDiT-XL, we will use two patch size $\{2, 4\}$. With $k=6$, majority of transformer block operates on only 64 image tokens while only last $6$ blocks operates on 256 image tokens, leading to significant GFLOPs reduction. Furthermore, we could train the model

\begin{table}[t]
\centering
\resizebox{\linewidth}{!}{
\begin{tabular}{lccccccc}
\toprule
Model & Epochs & GFLOPs$\downarrow$ & FID$\downarrow$ & sFID$\downarrow$ & IS$\uparrow$ & Pre$\uparrow$ & Rec$\uparrow$ \\
\midrule
ADM \cite{dhariwal2021diffusion}               & --     & --     & 10.94 & 6.02 & 100.98 & 0.69 & 0.63 \\
ADM-U             & --     & --     &  7.49 & 5.13 & 127.49 & 0.72 & 0.63 \\
% ADM-G             & --     & --     &  4.59 & 5.25 & 186.70 & 0.82 & 0.52 \\
ADM-G, ADM-U      & --     & --     &  3.94 & 6.14 & 215.84 & 0.83 & 0.53 \\
\midrule
CDM               & --     & --     &  4.88 & --   & 158.71 & --   & --  \\
\midrule
LDM-8  \cite{rombach2022high}            & --     & --     & 15.51 & --   &  79.03 & 0.65 & 0.63 \\
LDM-8-G           & --     & --     &  7.76 & --   & 209.52 & 0.84 & 0.35 \\
% LDM-4-G           & --     & --     &  3.95 & --   & 178.22 & 0.81 & 0.55 \\
LDM-4-G           & --     & --     &  3.60 & --   & 247.67 & \textbf{0.87} & 0.48 \\
\midrule
\multicolumn{8}{l}{\emph{Mamba and State Space Diffusion Model}} \\
\midrule
DiM-H/2 \cite{teng2024dim}          & --     & --     &  2.40 & --   & --     & --   & --  \\
% DiS-H/2           & --     & --     &  2.10 & 4.55 & 271.32 & 0.82 & 0.58 \\
DIFFUSSM-XL  \cite{yan2024diffusion}     & 515    & --     &  2.28 & 4.49 & 259.13 & 0.86 & 0.56  \\
DiMSUM-L/2-G  \cite{phung2024dimsum}    & 510    & --     &  2.11 & -- & --     & --   & 0.59  \\
\midrule
\multicolumn{8}{l}{\emph{Our Model and Baselines}} \\
\midrule
SiT-B/2     \cite{ma2024sit}        & 80  & 23.02  & 34.84 & 6.59  & 41.53 & 0.52 & 0.64  \\
DiT-B/2  \cite{peebles2023scalable}           & 80  & 23.02  & 43.47 & --    & --    & --   & --  \\
DiG-B/2  \cite{zhu2025dig}           & 80  & 17.07  & 39.50 & 8.50  & 37.21 & --   & --  \\
DiC-B   \cite{tian2025dic}            & 80  & 23.50  & 32.33 & --    & 48.72 & 0.51 & 0.63 \\
DiCo-B      \cite{ai2025dico}        & 80  & 16.88  & 27.20 & --    & 56.52 & \textbf{0.60} & 0.61 \\
MPDiT-B  (ours)           & 80  & \textbf{16.60} &\textbf{24.74} & \textbf{6.32}  & \textbf{57.40} & 0.58 & \textbf{0.65} \\
\midrule
SiT-XL/2    \cite{ma2024sit}        & 80  & 118.66 & 18.04 & 5.07  & 73.90 & 0.63 & 0.64 \\
DiT-XL/2     \cite{peebles2023scalable}       & 80  & 118.66 & 19.47 & --    & --    & --   & --  \\
DiG-XL/2     \cite{zhu2025dig}       & 80  & 89.40  & 18.53 & 6.06  & 68.53 & 0.63 & 0.64 \\
DiC-XL   \cite{tian2025dic}           & 80  & 116.1  & 13.11 & --    & 100.15& --   & --  \\
DiCo-XL   \cite{ai2025dico}          & 80  & 87.30  & 11.67 & --    & 100.42& \textbf{0.71} & 0.61 \\
MPDiT-XL  (ours)          & 80  & \textbf{59.30} & \textbf{9.92}  & \textbf{5.05} & \textbf{102.79} & 0.70 & \textbf{0.64} \\
\midrule
SiT-XL/2    \cite{ma2024sit}        & 1400 & 118.66 & 9.35  & 6.38  & 126.06 & 0.67 & 0.68 \\
DiT-XL/2     \cite{peebles2023scalable}       & 1400 & 118.66 & 9.62  & 6.85  & 121.50 & 0.67 & 0.67 \\
DiG-XL/2     \cite{zhu2025dig}       & 240  & 89.40  & 8.60  & 6.46  & \textbf{130.03} & 0.68 & 0.68 \\
MPDiT-XL (ours)           & 240  & \textbf{59.30} & \textbf{7.36}  & \textbf{5.09}  & 127.57 & \textbf{0.69} & \textbf{0.69} \\
\midrule
SiT-XL/2-G     \cite{ma2024sit}     & 1400 & 118.66 & 2.15  & 4.60  & 258.09 & 0.81 & 0.60 \\
DiT-XL/2-G   \cite{peebles2023scalable}       & 1400 & 118.66 & 2.27  & 4.60  & 278.24 & \textbf{0.83} & 0.57 \\
DiG-XL/2-G      \cite{zhu2025dig}    & 240  & 89.40  & 2.07  & 4.53 & \textbf{278.95} & 0.82 & 0.60 \\
MPDiT-XL-G  (ours)        & 240  & \textbf{59.30} & \textbf{2.05} & \textbf{4.51}  & 278.73 & 0.82 & \textbf{0.61} \\
\bottomrule
\end{tabular}
} 
\caption{Quantitative performance of MPDiT on ImageNet 256.}
\label{tab:imagenet256}
\end{table}

In the main experiments, the configuration of MPDiT is provided in \cref{tab:mpdit_config}. For the experiment on ImageNet $256 \times 256$ with two variants MPDiT-B and MPDiT-XL, we will use two patch size $\{2, 4\}$. With $k=6$, majority of transformer block operates on only 64 image tokens while only last $6$ blocks operates on 256 image tokens, leading to significant GFLOPs reduction. Furthermore, we could train the model on low memory GPUs with higher batch size (i.e 1024). In sample process, the throughput (sample/s) of MPDiT-XL is more than $2\times$ faster than DiT-XL/2 architecture.

As shown in \cref{tab:imagenet256}, MPDiT outperforms the baseline architectures in both generation quality and computational efficiency. 
Our XL configuration achieves a non-cfg FID of $7.36$ and a cfg FID of $2.05$ (with cfg scale $1.4$) after only 240 training epochs. 
In contrast, the SiT baseline requires 1400 epochs to reach a FID of $9.35$. 
These results indicate that MPDiT provides substantially improved training efficiency while maintaining high-fidelity generation. For the qualitative results, please refer to \cref{fig:teaser} and Appendix. 
%\quan{update result here and table}

\begin{table}[t]
\centering
\resizebox{\linewidth}{!}{
\begin{tabular}{lccccc}
\toprule
Model & $N$ & $k$ & Model Dim $D$ & GFLOPs $\downarrow$ & $\tfrac{\text{GFLOPs}_{\text{MPDiT}}}{\text{GFLOPs}_{\text{DiT}}} \downarrow$ \\
\midrule
MPDiT-B  & 12 & 6 & 768  & 16.6 & 72.1\% \\
%MPDiT-L  & 24 & 6 & 1024 & 39.5 & 48.8\% \\
MPDiT-XL & 28 & 6 & 1152 & 59.3 & 49.9\% \\
\bottomrule
\end{tabular}
} % end resizebox
\caption{Model configuration and computational cost of MPDiT.}
\label{tab:mpdit_config}
\end{table}
% \vspace{-2mm}

\subsection{Ablation Study} \label{sec:exp:ablate}
In this section, we perform ablation studies using the MPDiT-B configuration to verify the effectiveness of each architectural component. 
All ablation models are trained under the same settings described above, and we report 50k-FID using Euler sampling with 250 NFEs for evaluation. 

\begin{table}[t]
\centering
\scriptsize
\begin{tabular}{lcccc}
\toprule
Method & Params (M) & GFLOPs$\downarrow$ & FID$\downarrow$ \\
\midrule
DiT-B/2 & 130 .0 & 23.0  & 34.84 \\
+ Shared AdaIN $(\alpha, \gamma, \beta)$ & 90.3  & 22.9 & 35.31 \\
+ Multitokens Class $(m=16)$ & 101.9 & 24.3 & 28.56 \\
+ FNO Time Embedding & 101.2 & 24.3 & 24.52  \\
+ MPDiT $(k=6)$ \textbf{\textit{(default)}} & 104.8 & 16.6 & 24.74 \\
% - Skip Connection & -- & -- & 25.67 \\
\bottomrule
\end{tabular}
\caption{Ablation on MPDiT components. All models are trained for 80 epochs under the same settings. 
Note that in the second row (shared AdaIN), we still apply AdaIN to the combined class and time embeddings as original DiT.
}
\label{tab:efficiency_fid}
\end{table}

\begin{figure*}[t]
\centering

% --------------------- Row 1 ---------------------
\begin{minipage}[t]{0.45\linewidth}
\centering
\begin{table}[H]
\centering
\small
\begin{tabular}{lccc}
\toprule
Method & Params(M) & GFLOPs$\downarrow$ & FID$\downarrow$ \\
\midrule
\multicolumn{4}{c}{\emph{Configuration B}} \\
\midrule
DiT-B/2 $\dagger$     & 101.2 & 24.3 & 24.52 \\
MPDiT $k=4$ & 104.8 & 13.9 & 26.94 \\
MPDiT $k=6$ \textbf{\textit{(default)}} & 104.8 & 16.6 & 24.74 \\
MPDiT $k=8$ & 104.8 & 19.3 & 24.62 \\
\midrule
\multicolumn{4}{c}{\emph{Configuration XL}}  \\
\midrule
DiT-XL/2 $\dagger$   & 473.1 &125.5 & 9.22 \\
MPDiT $k=4$ & 481.2 & 53.2 & 11.11 \\
MPDiT $k=6$ \textbf{\textit{(default)}} & 481.2 & 59.3 & 9.92 \\
MPDiT $k=8$ & 481.2 & 65.4 & 9.73 \\
\bottomrule
\end{tabular}
\caption{Ablation on $k$ value for MPDiT. $\dagger$ means that DiT with Shared AdaIN, multitokens class and FNO time embedding.}
\label{tab:k_value}
\end{table}
\end{minipage}
\hfill
\begin{minipage}[t]{0.48\linewidth}
\centering
\begin{table}[H]
\centering
\small
\begin{tabular}{lccc}
\toprule
Method & Params(M) & GFLOPs$\downarrow$ & FID$\downarrow$ \\
\midrule
\multicolumn{4}{l}{\textit{Traditional Time Embedding}} \\
\midrule
$1\times$\texttt{Linear} & 104.9 & 16.6 & 28.98 \\
$2\times$\texttt{Linear} \textbf{\textit{(default)}} & 105.5 & 16.6 & 27.92 \\
$3\times$\texttt{Linear} & 106.1 & 16.6 & 29.71 \\
\midrule
\multicolumn{4}{l}{\textit{FNO Time Embedding}} \\
\midrule
$2\times$\texttt{MixedFNO} & 104.8 & 16.6 & 26.87 \\
$3\times$\texttt{MixedFNO} \textbf{\textit{(default)}} & 104.8 & 16.6 & 24.74 \\
$4\times$\texttt{MixedFNO} & 104.8 & 16.6 & 27.37 \\
\bottomrule
\end{tabular}
\caption{Ablation on the time embedding module. 
We vary the number of \texttt{Linear} layers and \texttt{MixedFNO} blocks. 
The traditional time embedding applies two \texttt{Linear} layers to sinusoidal time features, 
whereas the proposed FNO time embedding uses three \texttt{MixedFNO} blocks operating on grid-based time features.
}
\label{tab:time}
\end{table}
\end{minipage}

% --------------------- Row 2 ---------------------
\begin{minipage}[t]{0.45\linewidth}
\centering
\begin{table}[H]
\centering
\small
\begin{tabular}{lccc}
\toprule
Method & Params(M) & GFLOPs$\downarrow$ & FID$\downarrow$ \\
\midrule
$m=1$ & 93.3 & 15.3 & 32.31 \\
$m=4$ & 95.6 & 15.6 & 30.91 \\
$m=8$ & 98.7 & 15.9 & 28.12 \\
$m=16$ \textbf{\textit{(default)}} & 104.8 & 16.6 & 24.74 \\
$m=32$ & 117.1 & 18.0 & 24.47 \\
\bottomrule
\end{tabular}
\caption{Ablation on number of class tokens $m$}
\label{tab:class}
\end{table}
\end{minipage}
\hfill
\begin{minipage}[t]{0.50\linewidth}
\centering
\begin{table}[H]
\small
\centering
\begin{tabular}{lccc}
\toprule
Method & Params(M) & GFLOPs$\downarrow$ & FID$\downarrow$ \\
\midrule
\texttt{Linear} & 104.2 & 16.5 & 27.06 \\
\texttt{ConvTranpose} & 104.2 & 16.9 & 29.45 \\
\texttt{Linear+Linear} \textbf{\textit{(default)}} & 104.8 & 16.6 & 24.74 \\
\texttt{Linear+MLP(r=4)} & 109.0 & 17.8 & 24.85 \\
\texttt{Linear+Conv} & 104.8 & 16.6 & 25.68 \\
\bottomrule
\end{tabular}
\caption{Ablation on design of Upsample Block. $r$ is the mlp ratio.}
\label{tab:upsample}
\end{table}
\end{minipage}
\end{figure*}

In \cref{tab:efficiency_fid}, we analyze how each proposed component affects both model performance (FID) and efficiency (parameters and GFLOPs). 
We first apply the shared AdaIN strategy to the combined time and class embeddings, replacing the per-block AdaIN used in DiT. This change reduces the number of parameters from 130M to 90M, approximately a 30\% reduction, while keeping GFLOPs unchanged and increasing FID by only 0.4. 
This shows that local AdaIN layers can be safely removed in favor of a global shared AdaIN without causing significant degradation. Next, the proposed multi-token class embedding provides a substantial improvement, reducing FID by roughly 7 points. This suggests that a single class token is insufficient to encode class semantics, whereas multiple class tokens offer a richer representation and provide stronger supervision to the image tokens, resulting in faster convergence and better performance.
In addition, the FNO-based time embedding yields a further improvement of about 4 FID points. 
This indicates that FNO layers capture smoother and more informative temporal structure for flow matching, which aligns well with the underlying ODE formulation of diffusion and flow dynamics.
Overall, revisiting the design of the time and class embedding modules leads to nearly a 10-point FID improvement with only a small increase in computation (1.3 GFLOPs) and still reduces the parameter count by 30M. 
Finally, replacing the isotropic DiT architecture with our hierarchical MPDiT design further reduces GFLOPs from 23 to 16.6 while maintaining strong performance.

\paragraph{Ablation on $k$ number of last blocks. }\cref{tab:k_value} shows that MPDiT requires only a relatively small number of fine-resolution blocks to maintain strong performance. For both the B and XL configurations, using $k = 6$ results in less than a 1 point drop in FID compared to DiT isotropic design while providing substantial efficiency gains. 
\vspace{-2mm}

\paragraph{Ablation on time embedding design. }In the conventional sinusoidal linear time embedding, we experiment with reducing and increasing the number of \texttt{Linear} layers to assess whether deeper MLPs can capture richer temporal information. 
We also vary the number of \texttt{MixedFNO} layers in our proposed design. 
The results show that the FNO-based time embedding consistently improves FID by approximately 3 points compared to the standard embedding. 
Increasing the number of layers in either time embedding design tends to degrade performance, due to gradient instability introduced by deeper time-embedding module.
\vspace{-2mm}

\paragraph{Ablation on number of class tokens $m$. }\cref{tab:class} shows that using 16 class tokens provides a substantial improvement in FID while keeping GFLOPs unchanged. 
Increasing the number of class tokens to 32 further raises the computational cost but yields only a marginal additional improvement in performance. This finding highlights the potential for better exploiting label signals to maximize the information available during training, thereby improving model performance and accelerating convergence.
\vspace{-2mm}

\paragraph{Ablation on design of Upsample Block. }In \cref{tab:upsample}, replacing the \texttt{ConvTranspose} layer with a \texttt{Linear} layer for expanding coarse global tokens into finer local tokens yields noticeably better performance. 
Based on this observation, we adopt the \texttt{Linear} upsampling strategy in MPDiT. 
We further evaluate several refinement mechanisms applied after upsampling. 
The results show that adding a single additional \texttt{Linear} layer provides the best performance, 
while using more complex modules such as an \texttt{MLP} or a \texttt{Conv} block leads to degraded results. 
\vspace{-2mm}

%% file: sec/5_conclusion.tex
\section{Conclusion}\label{sec:con}
In this paper, we revisit the design of the time and class embedding and show that simple modifications can yield substantial performance gains, improving FID by 10 points. 
In addition, our proposed global-local MPDiT architecture, combined with a carefully designed upsampling module, reduces GFLOPs by up to 50\% while maintaining the original performance. 
This results in notable improvements in training efficiency, memory usage, and sampling speed.

\textbf{Limitations.} Although MPDiT demonstrates strong efficiency in image generation, extending the architecture to large-scale settings such as text-to-image models (e.g., SDv3, Flux) and text-to-video models (e.g., Sora models) remains an open direction. 
While these tasks appear promising for further exploration, they require substantial computational resources. 
%Due to the limited compute available in academic environments, we leave large-scale experiments to future work.

%% file: sec/X_suppl.tex
\clearpage
\setcounter{page}{1}
\maketitlesupplementary
In the supplementary, we first present the results in MPDiT in \cref{sec:imagenet512}. \cref{sec:converge} summaries the training time and convergence of our method against the baseline DiT/SiT. Detail of FNO time embedding implementation is provided in \cref{sec:fno_details}. Finally, we include more qualitative results in \cref{sec:qual}. 

\begin{figure*}[th!]
    \centering
    \includegraphics[width=1.0\linewidth]{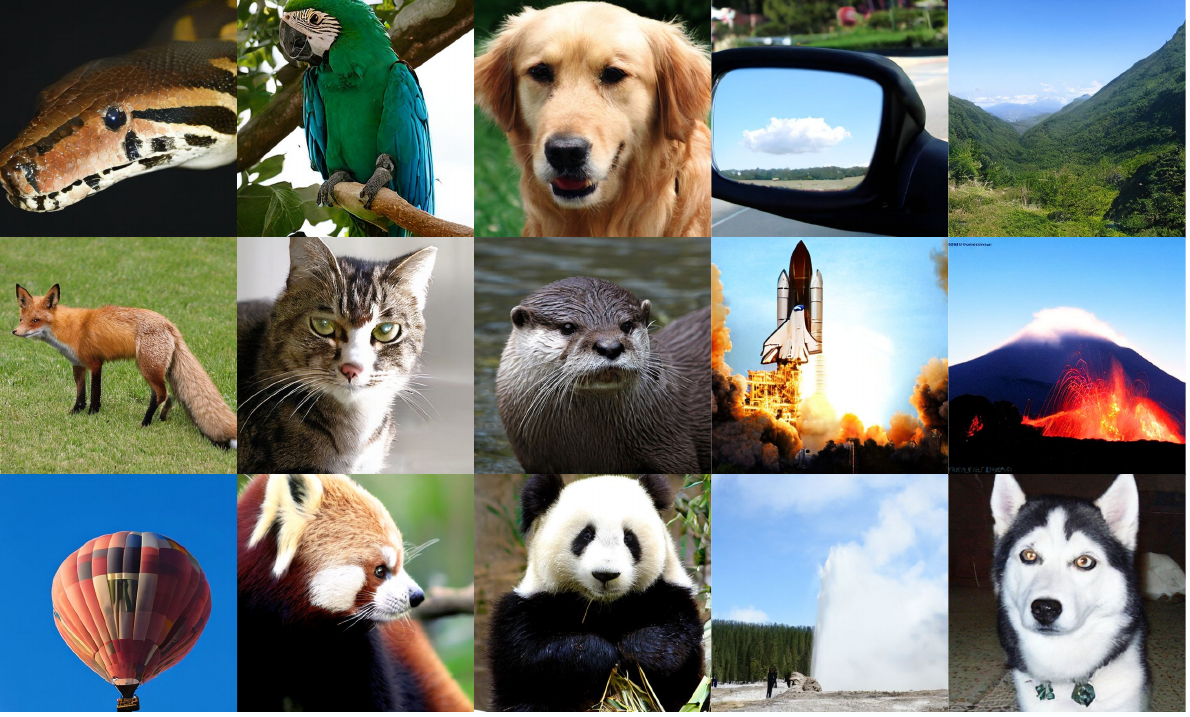}
    \caption{Qualitative Result of Imagenet 512 with cfg=4}
    \label{fig:512}
\end{figure*}

\section{ImageNet 512 results}
\label{sec:imagenet512}

\begin{table}[H]
\centering
\scriptsize
\resizebox{0.95\linewidth}{!}{
\begin{tabular}{l|c c|c c c}
\toprule
 Model & Params(M) & GFLOPs & FID $\downarrow$ & Prec $\uparrow$ & Rec $\uparrow$ \\
\midrule
 BigGAN \cite{brock2018large} & 160 & - & 8.43 & \textbf{0.88} & 0.29 \\
 StyleGAN-XL \cite{sauer2022stylegan} & 166 & - & 2.41 & 0.77 & 0.52 \\
\midrule
 MaskGIT \cite{chang2022maskgit} & 227 & - & 7.32 & 0.78 & 0.50 \\
\midrule
 VQ-GAN \cite{esser2021taming} & 227 & - & 26.52 & 0.73 & 0.31 \\
 VAR-d36-s \cite{tian2024visual} & 2300 & - & 2.63 & - & - \\
\midrule
 ADM-U \cite{dhariwal2021diffusion} & 731 & 2813 & 3.85 & 0.84 & 0.53 \\
 U-ViT-L/4 \cite{bao2023all} & 287 & 76.5 & 4.67 & 0.87 & 0.45 \\
 U-ViT-H/4 \cite{bao2023all} & 501 & 133.3 & 4.03 & 0.84 & 0.48 \\
 Simple Diff \cite{hoogeboom2023simple} & 2000 & - & 4.53 & - & - \\
 VDM++ \cite{kingma2023understanding} & 2000 & - & 2.65 & - & - \\
 DiT-XL/2 \cite{peebles2023scalable} & 675 & 524.7 & 3.04 & 0.84 & 0.54 \\
 SiT-XL \cite{ma2024sit} & 675 & 524.7 & 2.62 & 0.84 & \textbf{0.57} \\
 DiM-H \cite{teng2024dim} & 860 & 708 & 3.78 & - & - \\
 DiffuSSM-XL \cite{yan2024diffusion} & 673 & 1066.2 & 3.41 & 0.85 & 0.49 \\
 DiCo-XL \cite{ai2025dico} & 701 & 349.8 & 2.53 & 0.83 & 0.56 \\
 MPDiT-XL (ours) & 482 & 228.4 & \textbf{2.47} & 0.83 & 0.56 \\
\bottomrule
\end{tabular}
}
\caption{Quantitative results of ImageNet 512 with MPDiT$_{k=6}$}
\label{tab:512_performance}
\end{table}

\begin{table}[H]
\centering
\scriptsize
\begin{tabular}{lcccc}
\toprule
Method &Epoch & Params(M)  & GFLOPs$\downarrow$ & FID$\downarrow$ \\
\midrule
\multicolumn{5}{c}{\emph{No cfg}}  \\
\midrule
DiT-XL/2   &600 & 675 &524.7 & 11.93 \\
MPDiT$_{\{22, 6\}}$ &120 & 482 & 228.4 & 9.24 \\
% MPDiT$_{\{14, 8, 6\}}$ &120 & 491 & 180.4 & - \\
MPDiT$_{\{18, 6, 4\}}$ &120 & 491 & 138.2 & 11.77 \\
\midrule
\multicolumn{5}{c}{\emph{With cfg}}  \\
\midrule
DiT-XL/2   &600 & 675 &524.7 & 3.04 \\
SiT-XL/2   &600 & 675 &524.7 & 2.62 \\
MPDiT$_{\{22, 6\}}$ &120 & 482 & 228.4 & 2.47 \\
% MPDiT$_{\{14, 8, 6\}}$ &120 & 491 & 180.4 & - \\
MPDiT$_{\{18, 6, 4\}}$ &120 & 491 & 138.2 & 3.13 \\
\bottomrule
\end{tabular}
\caption{Performace of different MPDiT-XL variants on ImageNet 512}
\label{tab:512_variant}
\end{table}

\paragraph{Training Details:} All $512^2$ experiments are conducted on a single node A100 (40GB) with a total batch size of $256$, trained for $120$ epochs (approximately $600\text{K}$ iterations). Notably, MPDiT reduces memory consumption, allowing the full batch size of $256$ to fit within a single node of A100 40GB GPU.

\paragraph{Sampling Details:} For sampling process, we adopt Euler sampling with $250$ NFEs. For guided sampling, we use cfg scale $1.375$. We follow the evaluation protocol from \cite{dhariwal2021diffusion} to sample $50,000$ images and compute the evaluation metrics with $1000$ provided reference images.

\paragraph{Main Results:} As shown in \cref{tab:512_performance}, MPDiT-XL$_{k=6}$, which applies patch size $4$ to the first $22$ transformer blocks and patch size $2$ to the last $6$ blocks, achieves an FID of $2.47$, outperforming all baselines. Remarkably, it reaches this performance within only $120$ training epochs and requires just $228.4$ GFLOPs, corresponding to merely $43.5\%$ of the GFLOPs of the DiT and SiT baselines. The non-cherrypicked qualitative results of MPDiT-XL$_{k=6}$ is shown in \cref{fig:512}.

\paragraph{Performance of different variant MPDiT-XL:} We further evaluate several MPDiT-XL variants. \textbf{MPDiT$_{\{22,6\}}$} (equivalent to MPDiT-XL$_{k=6}$) denotes a two-stage configuration in which the first $22$ Transformer blocks use patch size $4$ and the final $6$ blocks use patch size $2$. \textbf{MPDiT$_{\{18,6,4\}}$} extends this to a three-level hierarchy: the first $18$ blocks use patch size $8$, the next $6$ blocks use patch size $4$, and the final $4$ blocks use patch size $2$. As shown in \cref{tab:512_variant}, MPDiT$_{\{22,6\}}$ already surpasses the DiT/SiT baselines after only $120$ training epochs while requiring just $228.4$ GFLOPs. To further reduce computational cost, we explore MPDiT$_{\{18,6,4\}}$, which achieves a strong trade-off between performance and efficiency. Its non-guided version outperforms DiT-XL/2 with an FID of \textbf{11.77}, despite being trained for only $120$ epochs. With classifier-free guidance, it attains an FID of \textbf{3.13}, slightly behind DiT-XL/2. However, it is important to highlight that MPDiT$_{\{18,6,4\}}$ uses only $\sim 26\%$ of the GFLOPs of DiT/SiT and only is trained for 120 epochs, suggesting substantial room for improvement with longer training. Finally, note that SiT-XL/2 is evaluated using \textbf{SDE sampling} (FID $2.62$), whereas all MPDiT results are obtained using \textbf{ODE sampling}. Since SDE typically yields better sample quality than ODE, the comparison in \cref{tab:512_variant} is not strictly apple-to-apple.

\section{Convergence \& Efficiency Analysis}
\label{sec:converge}

\paragraph{Imagenet 256:} \cref{tab:imagenet256} shows that MPDiT-XL achieves an FID of $2.05$ with only $240$ training epochs and $59.3$ GFLOPs. This corresponds to just $17.2\%$ of the training epochs and $49.9\%$ of the per-iteration GFLOPs of the DiT baseline. Overall, the total training compute of MPDiT-XL amounts to only $8.8\%$ of that required by DiT and SiT, indicating a convergence that is approximately $11.36\times$ faster.

\paragraph{Imagenet 512:} \cref{tab:512_performance} demonstrates that MPDiT-XL achieves an FID of $2.47$, outperforming all compared methods while training for only $120$ epochs with $228.4$ GFLOPs. This corresponds to merely $20\%$ of the training epochs and $43.5\%$ of the per-iteration GFLOPs of DiT/SiT, resulting in a total training compute of only $8.7\%$. Consequently, MPDiT converges approximately $11.5\times$ faster than DiT/SiT.

\paragraph{Inference Time:} Under the same GPU and number of function evaluations (NFEs), MPDiT achieves more than $2\times$ faster sampling compared to the DiT and SiT baselines, while also consuming less memory. This improved efficiency enables MPDiT to sample effectively across a wider range of GPU devices.

\paragraph{Training Memory Consumption:} During training, under identical settings, MPDiT enables significantly larger batch sizes than baseline DiT/SiT. For $256\times256$ resolution, we can fit a total batch size of $1024$ on a single node with $8$ A100 (40GB) GPUs, which is infeasible for DiT/SiT due to their higher GFLOPs. For $512\times512$ resolution, MPDiT similarly fits a batch size of $256$ on the same hardware, while DiT/SiT cannot. These results demonstrate that MPDiT allows efficient diffusion/flow matching training without requiring multi-node clusters or higher-memory GPUs.

\section{FNO Time Embedding Details}
\label{sec:fno_details}
The detail implementation of FNO time embedding is provided in \cref{algo:1} and \cref{algo:2}. Note that in \cref{algo:1}, we use 1D grid as the time features. We have tried to replace 1D grid time feature with cos-sin sinusoidal time feature like in traditional time embedding and find out the model unable to converge.

\begin{algorithm*}[ht!]
\caption{PyTorch code of FNO time embedding}
\vspace{2mm}
% \resizebox{\textwidth}{!}{
% \begin{minipage}{.7\linewidth}
\hrule
\vspace{1mm}
\begin{lstlisting}[
    language=Python,
    basicstyle=\ttfamily\small,
    commentstyle=\color{green!40!black},
    keywordstyle=\color{blue!60!black},
    showstringspaces=false,
    columns=flexible
]
import torch
import torch.nn.functional as F

class FNOTimestepEmbedder(nn.Module):
    def __init__(self, hidden_size, modes=16, width=32):
        super().__init__()
        self.modes = modes
        self.width = width

        # lift scalar time to width-dim signal
        self.fc0 = nn.Linear(1, self.width)

        # spectral convolution layers
        self.conv0 = SpectralConv1d(self.width, self.width, self.modes)
        self.conv1 = SpectralConv1d(self.width, self.width, self.modes)
        self.conv2 = SpectralConv1d(self.width, self.width, self.modes)

        # local 1x1 convs
        self.w0 = nn.Conv1d(self.width, self.width, 1)
        self.w1 = nn.Conv1d(self.width, self.width, 1)
        self.w2 = nn.Conv1d(self.width, self.width, 1)

        # projection to model dimension
        self.fc1 = nn.Linear(self.width, hidden_size)

    def forward(self, t):
        B = t.shape[0]
        t = t.unsqueeze(-1)                       # (B, 1)

        # build 1D grid and center around timestep
        grid = torch.linspace(-1, 1, 32, device=t.device)
        grid = grid.unsqueeze(0).expand(B, -1)    # (B, 32)
        grid = grid + t                           # (B, 32)

        # lift to width dimension
        x = self.fc0(grid.unsqueeze(-1))          # (B, 32, W)
        x = x.permute(0, 2, 1)                    # (B, W, 32)

        # three FNO blocks (spectral + local conv)
        x = F.gelu(self.conv0(x) + self.w0(x))
        x = F.gelu(self.conv1(x) + self.w1(x))
        x = self.conv2(x) + self.w2(x)

        # average pool and project to model dim
        x = x.mean(dim=-1)                        # (B, W)
        x = self.fc1(x)
        return x
\end{lstlisting}
\vspace{1mm}
\hrule
\label{algo:1}
\end{algorithm*}

\begin{algorithm*}[ht!]
\caption{PyTorch code of 1D spectral convolution}
\vspace{2mm}
% \begin{minipage}{1.0\linewidth}
\hrule
\vspace{1mm}
\begin{lstlisting}[
    language=Python,
    basicstyle=\ttfamily\small,
    commentstyle=\color{green!40!black},
    keywordstyle=\color{blue!60!black},
    showstringspaces=false,
    columns=flexible
]
import torch

class SpectralConv1d(nn.Module):
    def __init__(self, in_channels, out_channels, modes):
        super().__init__()
        self.in_channels = in_channels
        self.out_channels = out_channels
        self.modes = modes

        # Fourier weights (real and imaginary parts)
        self.weights_real = nn.Parameter(
            torch.randn(in_channels, out_channels, modes)
        )
        self.weights_imag = nn.Parameter(
            torch.randn(in_channels, out_channels, modes)
        )

    def forward(self, x):
        # store dtype and switch to float32 for FFT
        dtype = x.dtype
        x_fp32 = x.float()

        # compute FFT
        x_ft = torch.fft.rfft(x_fp32, dim=-1)
        xr, xi = x_ft.real, x_ft.imag

        # allocate output spectrum
        B, C_out = x.shape[0], self.out_channels
        G = x.shape[-1] // 2 + 1
        out_r = torch.zeros(B, C_out, G, device=x.device)
        out_i = torch.zeros(B, C_out, G, device=x.device)

        # complex multiplication on first modes
        if self.modes <= xr.shape[-1]:
            xr_m = xr[:, :, :self.modes]
            xi_m = xi[:, :, :self.modes]

            wr = self.weights_real.float()
            wi = self.weights_imag.float()

            real = torch.einsum("bim,iom->bom", xr_m, wr) - \
                   torch.einsum("bim,iom->bom", xi_m, wi)
            imag = torch.einsum("bim,iom->bom", xr_m, wi) + \
                   torch.einsum("bim,iom->bom", xi_m, wr)

            out_r[:, :, :self.modes] = real
            out_i[:, :, :self.modes] = imag

        # inverse FFT to return to spatial domain
        out_ft = torch.complex(out_r, out_i)
        y = torch.fft.irfft(out_ft, n=x.shape[-1], dim=-1)

        # convert back to original precision
        return y.to(dtype)
\end{lstlisting}
\vspace{1mm}
\hrule
\label{algo:2}
\end{algorithm*}

% \section{Discuss About Class Embedding}
% \label{sec:class_embedding}

% discuss how it is related to REPA

% also add some exp on REPA if possible

\section{More Qualitative Results}\label{sec:qual}
\begin{figure*}[t]
    \begin{minipage}[t]{0.49\linewidth}
        \centering 
        \includegraphics[width=0.8\linewidth]{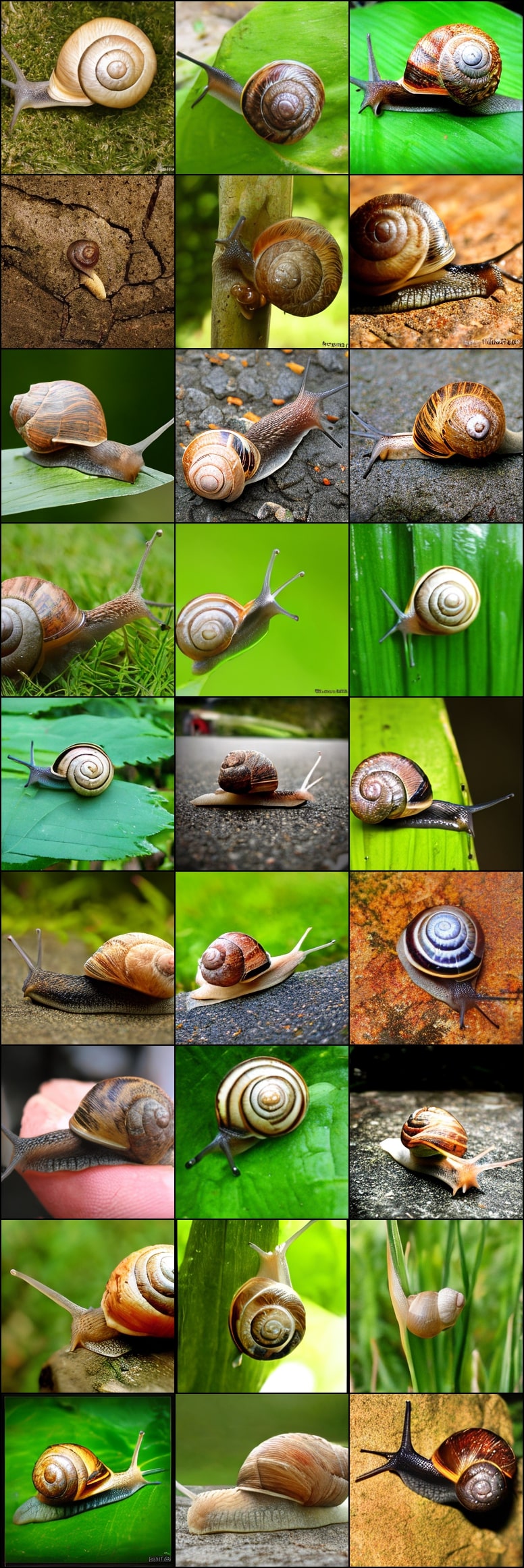}
        \caption{Qualitative images of class 113 "snail"}
        \label{fig:1}
    \end{minipage}
    \hfill
    \begin{minipage}[t]{0.49\linewidth}
        \centering
        \includegraphics[width=0.8\linewidth]{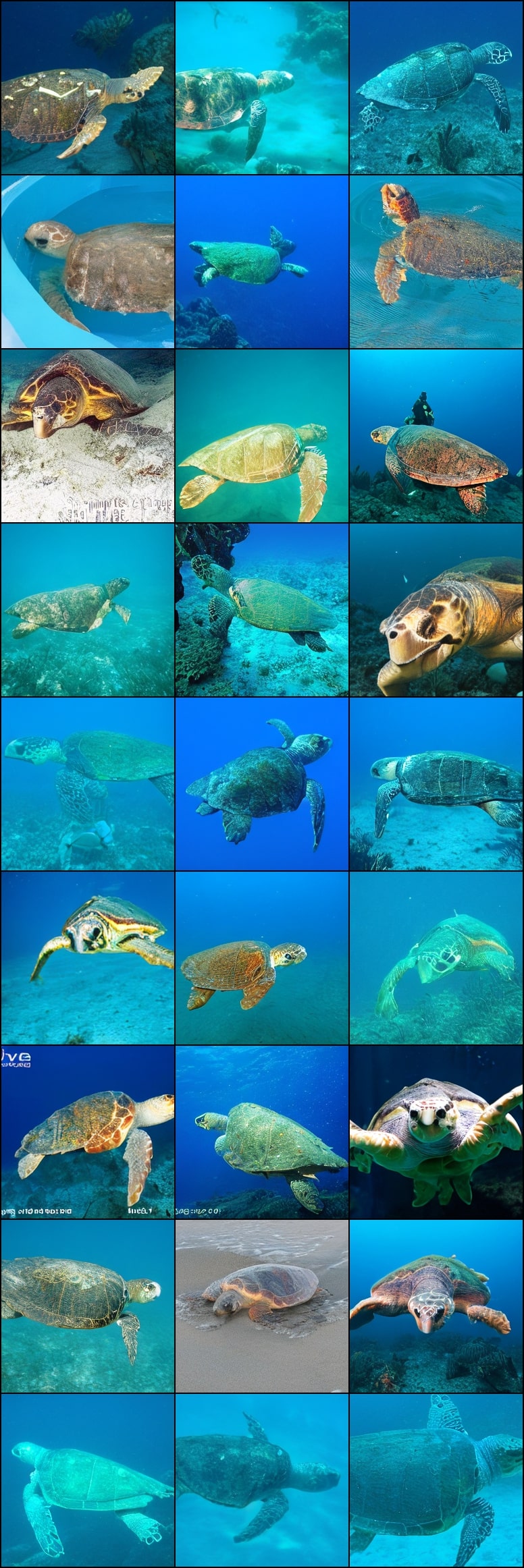}
        \caption{Qualitative images of class 33 "loggerhead, loggerhead turtle, Caretta caretta"}
        \label{fig:2}
    \end{minipage}
\end{figure*}

\begin{figure*}[t]
    \begin{minipage}[t]{0.49\linewidth}
        \centering 
        \includegraphics[width=0.8\linewidth]{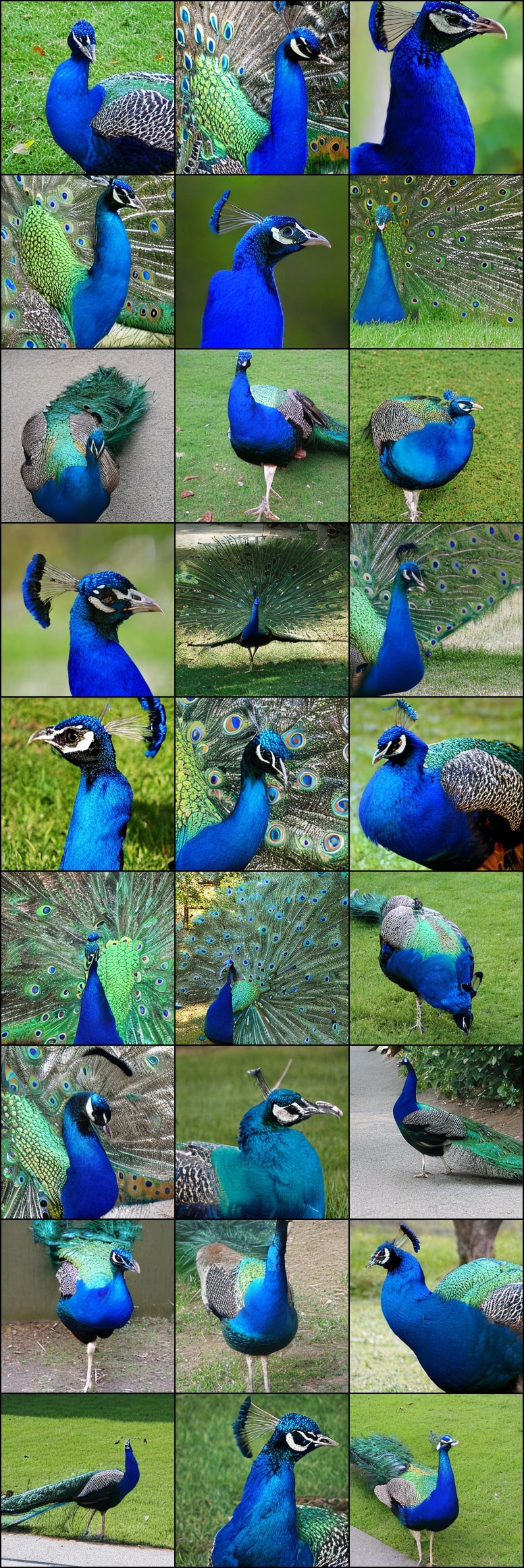}
        \caption{Qualitative images of class 84 "peacock"}
        \label{fig:3}
    \end{minipage}
    \hfill
    \begin{minipage}[t]{0.49\linewidth}
        \centering
        \includegraphics[width=0.8\linewidth]{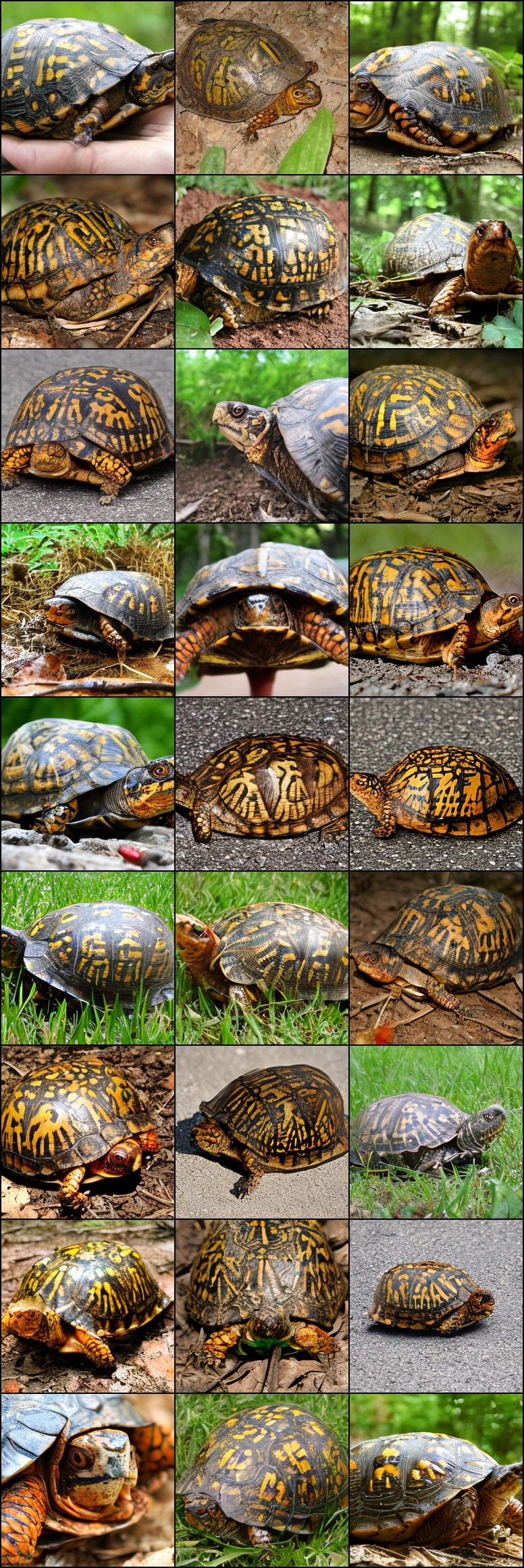}
        \caption{Qualitative images of class 37 "box turtle, box tortoise"}
        \label{fig:4}
    \end{minipage}
\end{figure*}

\begin{figure*}[t]
    \begin{minipage}[t]{0.49\linewidth}
        \centering 
        \includegraphics[width=0.8\linewidth]{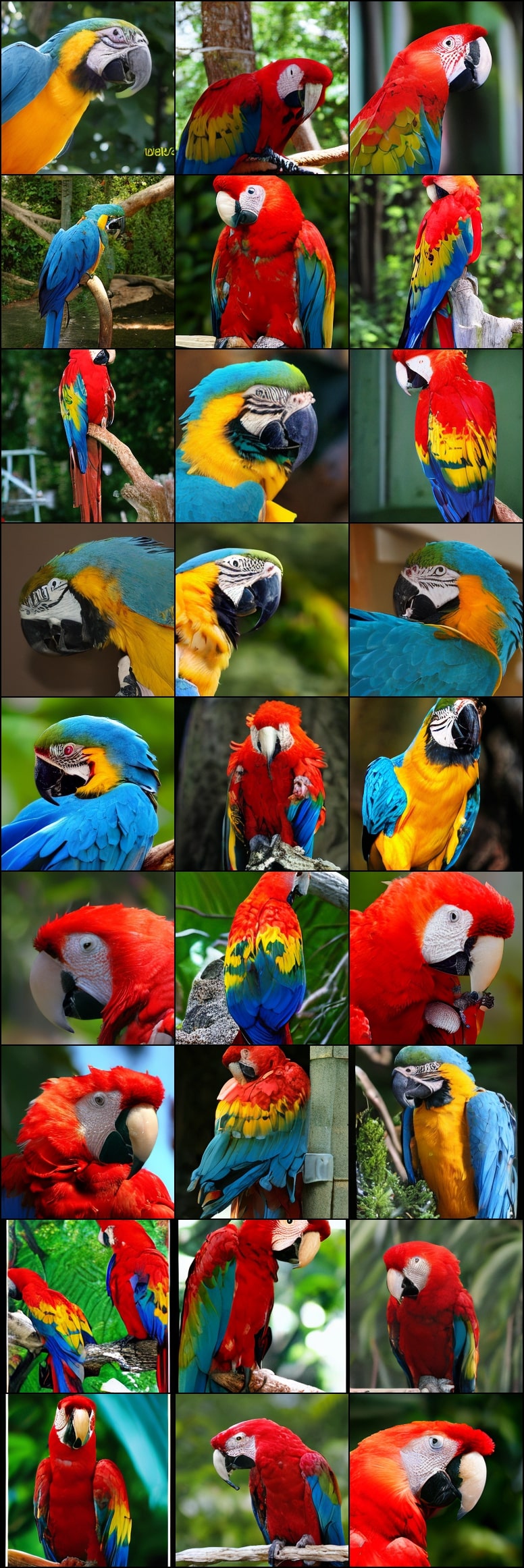}
        \caption{Qualitative images of class 88 "macaw"}
        \label{fig:5}
    \end{minipage}
    \hfill
    \begin{minipage}[t]{0.49\linewidth}
        \centering
        \includegraphics[width=0.8\linewidth]{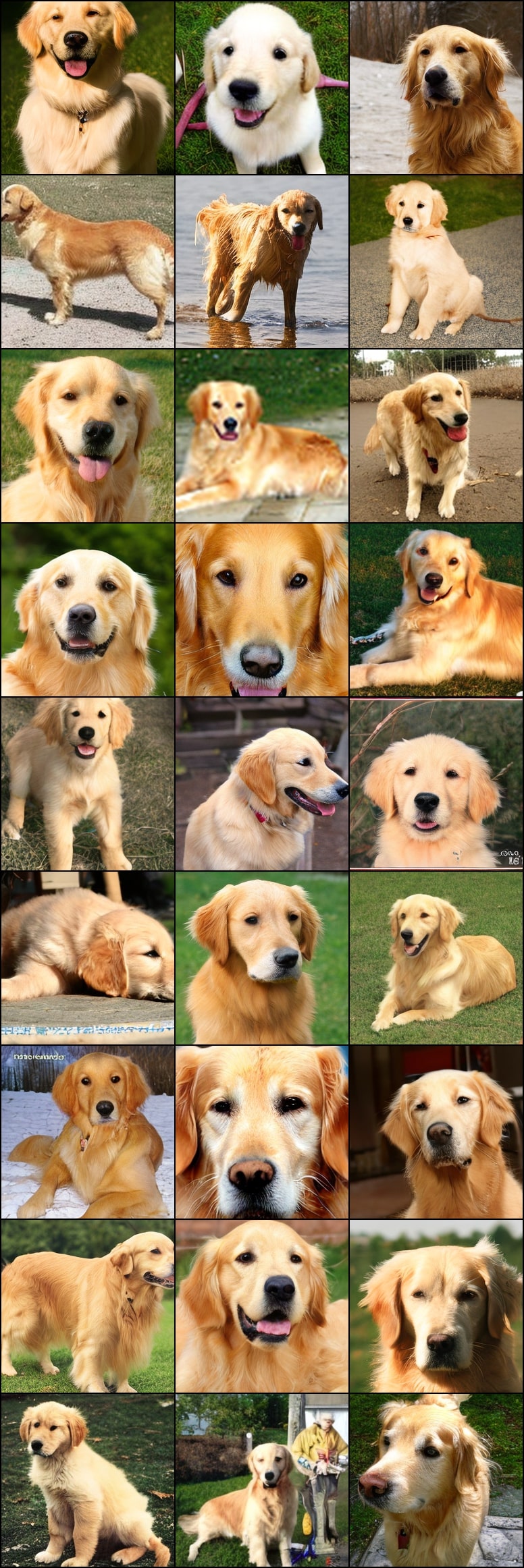}
        \caption{Qualitative images of class 207 "golden retriever"}
        \label{fig:6}
    \end{minipage}
\end{figure*}

\begin{figure*}[t]
    \begin{minipage}[t]{0.49\linewidth}
        \centering 
        \includegraphics[width=0.8\linewidth]{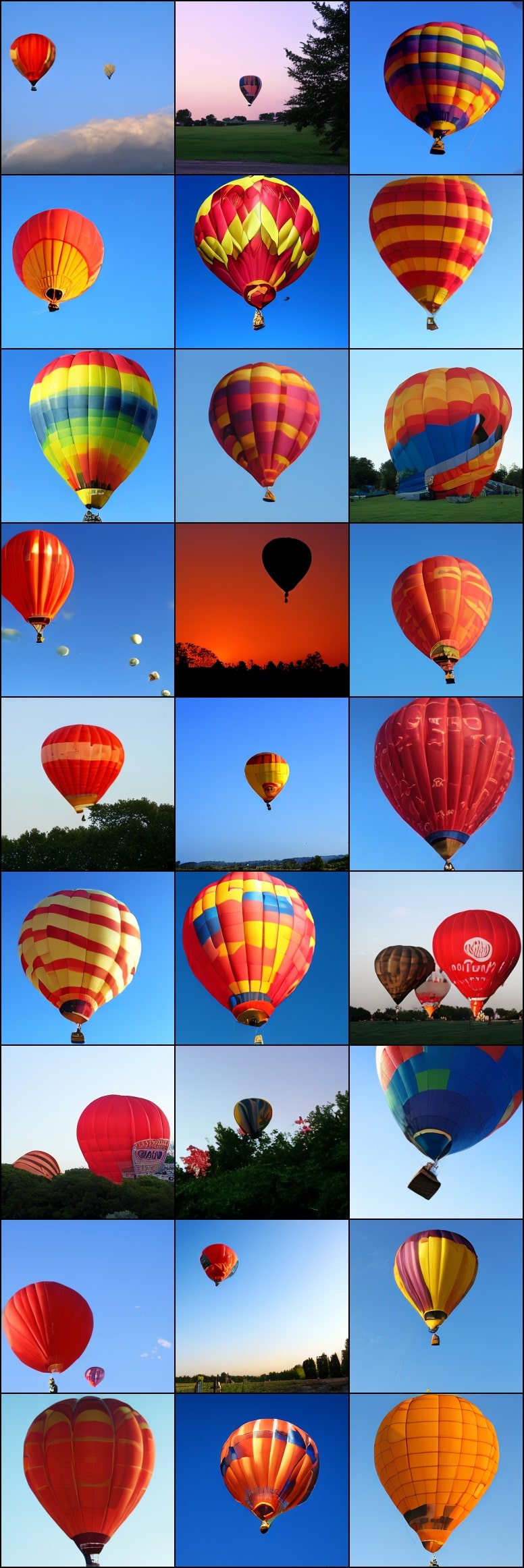}
        \caption{Qualitative images of class 417 "balloon"}
        \label{fig:7}
    \end{minipage}
    \hfill
    \begin{minipage}[t]{0.49\linewidth}
        \centering
        \includegraphics[width=0.8\linewidth]{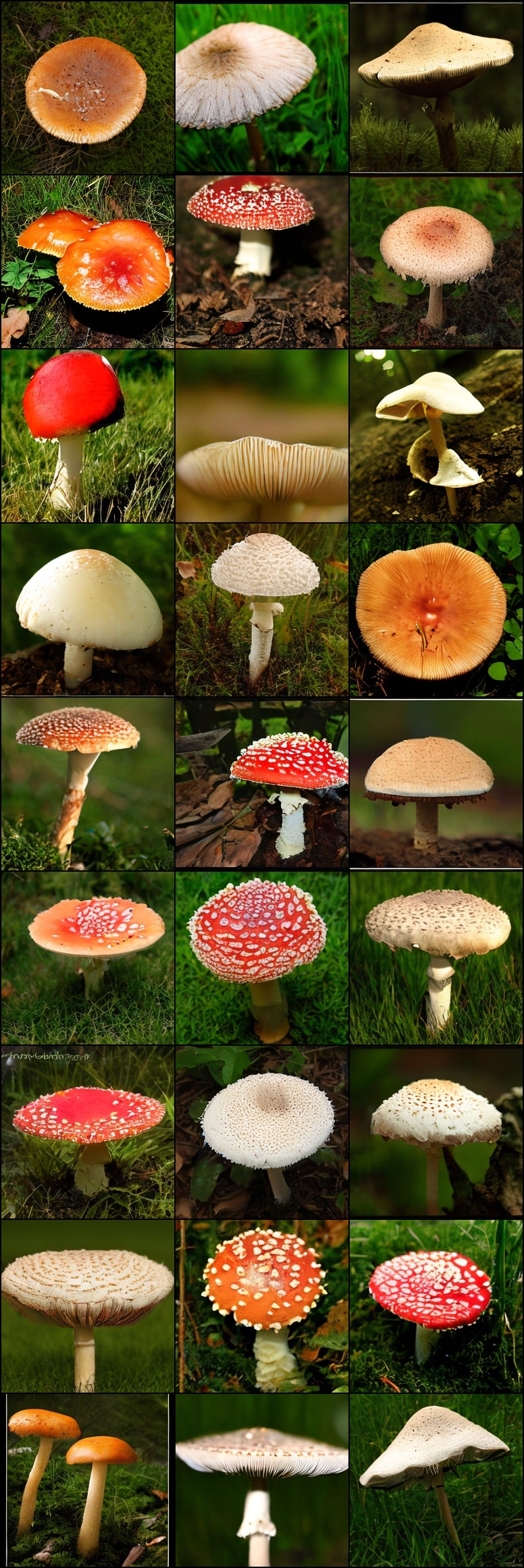}
        \caption{Qualitative images of class 947 "mushroom"}
        \label{fig:8}
    \end{minipage}
\end{figure*}

\begin{figure*}[t]
    \begin{minipage}[t]{0.49\linewidth}
        \centering 
        \includegraphics[width=0.8\linewidth]{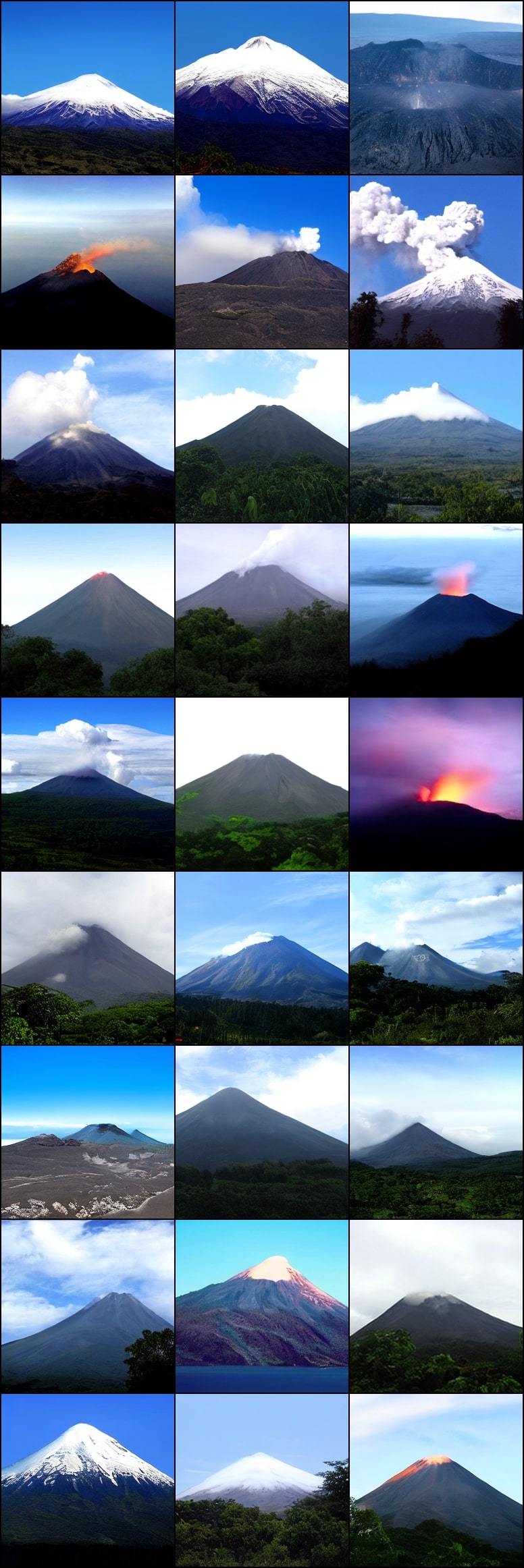}
        \caption{Qualitative images of class 980 "volcano"}
        \label{fig:9}
    \end{minipage}
    \hfill
    \begin{minipage}[t]{0.49\linewidth}
        \centering
        \includegraphics[width=0.8\linewidth]{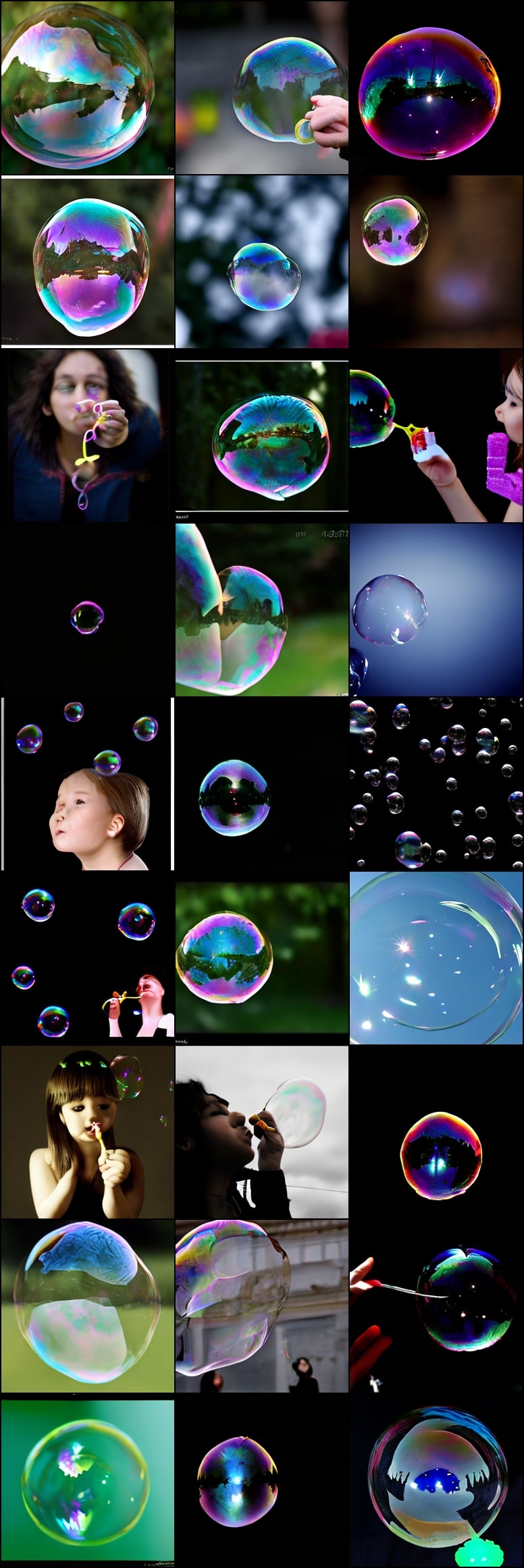}
        \caption{Qualitative images of class 971 "bubble"}
        \label{fig:10}
    \end{minipage}
\end{figure*}

More qualitative results are shown in \cref{fig:1}, \cref{fig:2}, \cref{fig:3}, \cref{fig:4}, \cref{fig:5}, \cref{fig:6}, \cref{fig:7}, \cref{fig:8}, \cref{fig:9} and \cref{fig:10}. All images are non-cherry pick and is sampled with Euler 250 steps and cfg scale is $4$.

